\pdfoutput=1

\documentclass[11pt]{article}

\usepackage[preprint]{acl}

\usepackage{times}
\usepackage{latexsym}

\usepackage[T1]{fontenc}

\usepackage[utf8]{inputenc}

\usepackage{microtype}

\usepackage{inconsolata}

\usepackage{graphicx}

\usepackage{inconsolata}
\usepackage{amsmath}  
\usepackage{graphicx}  
\usepackage{booktabs} 
\usepackage{algorithm}
\usepackage{algorithmic}
\usepackage{multirow}

\usepackage{enumitem}

\newcommand\blfootnote[1]{%
  \begingroup
  \renewcommand\thefootnote{}\footnote{#1}%
  \addtocounter{footnote}{-1}%
  \endgroup
}
\title{LOCUS:\textbf{A System and Method for Low-Cost Customization for Universal Specialization}}

\author{
Dhanasekar Sundararaman$^*$, \
Keying Li$^*$, \
Wayne Xiong, \
Aashna Garg \\
Microsoft \\
\texttt{{\{dhanasekars, keyli\}}@microsoft.com}
}

\begin{document}
\maketitle
\begin{abstract}

We present LOCUS (LOw‐cost Customization for Universal Specialization), a pipeline that consumes few-shot data to streamline the construction and training of NLP models through targeted retrieval, synthetic data generation, and parameter‐efficient tuning. With only a small number of labeled examples, LOCUS discovers pertinent data in a broad repository, synthesizes additional training samples via in‐context data generation, and fine‐tunes models using either full or low‐rank (LoRA) parameter adaptation. Our approach targets named entity recognition (NER) and text classification (TC)  benchmarks, consistently outperforming strong baselines (including GPT‐4o) while substantially lowering costs and model sizes. Our resultant memory-optimized models retain 99\% of fully fine‐tuned accuracy while using barely 5\% of the memory footprint, also beating GPT‐4o on several benchmarks with less than 1\% of its parameters.
\end{abstract}

\section{Introduction}

\blfootnote{$^*$equal contribution.}
Large Language Models (LLMs) \cite{touvron2023llama, team2023gemini, achiam2023gpt} have demonstrated remarkable versatility in zero-shot or few-shot prompting for many NLP tasks \cite{brown2020language}. Yet, relying on such models for real-world applications can be expensive \cite{xia2024understanding} and risky \cite{bommasani2021opportunities, huang2023survey}. The user often pays high inference costs or must accept unpredictable performance under domain shifts. Moreover, some deployments require on-premises (private) solutions \cite{kwon2024slm} or smaller model sizes for latency constraints \cite{irugalbandara2024scaling}. These practical requirements have led to a growing interest in building \emph{custom} or \emph{specialized} models that closely mirror LLM performance but with cheaper inference and tighter domain adaptation.

Named Entity Recognition (NER) and Text Classification (TC) are fundamental tasks in NLP that share many of the same challenges when label definitions shift or data is highly specialized. Traditional NER and TC often operates with standardized label sets and benefits from readily available datasets \cite{li2020survey}. Models such as GPT-4 and GPT-4o excel in zero-shot scenarios, but they may struggle when label definitions evolve or when data is highly customized \cite{wang2023gpt}. The emphasis of LOCUS lies on customizing NER and TC under dynamic labeling requirements and minimal supervision. By combining selective data retrieval, targeted synthetic expansions, and parameter-efficient updates, we build specialized models that is not only memory efficient but also outperforms LLMs including GPT-4o while addressing real-world needs for adaptability and efficiency.

Constructing high-quality training data for a custom NER or TC model, however, can be cumbersome. Collecting domain-specific labels manually is expensive, while purely synthetic data generation may produce low-diversity training corpora. \textbf{LOCUS} retrieves data at the \textit{sentence level} rather than extracting entire datasets, enabling more precise and granular selection of examples for new tasks. This approach supports the development of \textbf{few-shot customizable models} by allowing fine-grained data acquisition from multiple sources, which can be particularly valuable when users have minimal or specialized labeling requirements. 


\paragraph{Our contributions:} We introduce \textbf{LOCUS}—LOw‐cost Customization for Universal Specialization—a streamlined pipeline for lightweight dataset construction and customized model training. With minimal user effort, LOCUS carries out three core steps: \emph{Data Retrieval}, \emph{Synthetic Generation},  and \emph{Parameter-Efficient Fine-Tuning}.

\noindent We highlight key aspects of LOCUS:

\begin{itemize}
\item \textbf{Superior performance on NER and TC using synthetic data} Merging retrieved real-world examples with our synthetic expansions consistently outperforms state-of-the-art LLMs on several NER and TC datasets. Through comprehensive ablations, we find that LOCUS$_\text{mini}$ through LoRA adapters achieve accuracy within 1–2\% of full fine-tuning while cutting memory usage by as much as 90\%.
\item \textbf{Comparison with state-of-the-art.} LOCUS surpasses in-context GPT-4o performance on many standard benchmarks while using under 1\% of GPT’s overall parameters. LOCUS$_{mini}$'s 5M parameter model uses parameters that are an order of magnitude lower than other models with billions of parameters.
\end{itemize}

\begin{table*}[hbt!]
\centering
\begin{tabular}{l c c c c c c c c c}
\toprule
\textbf{Model} & \textbf{\textit{params}} & \multicolumn{2}{c}{\textbf{MIT}} & \multicolumn{5}{c}{\textbf{CrossNER}} & \multicolumn{1}{c}{\textbf{AVG}} \\
\cmidrule(lr){3-4}\cmidrule(lr){5-9}
& & \textbf{Movie} & \textbf{Res.} & \textbf{AI} & \textbf{Lit.} & \textbf{Music} & \textbf{Politics} & \textbf{Science} & \\
\midrule
\emph{zero-shot} & & & & & & & & & \\
Vicuna-13B & 13B & 0.9 & 0.4 & 22.7 & 22.7 & 26.6 & 27.0 & 22.0 & 17.5 \\
InstructUIE & 11B & 63.0 & 21.0 & 49.0 & 47.2 & 53.2 & 48.1 & 49.2 & 47.2 \\
GPT-3.5-turbo & 175B & 5.3 & 32.8 & 52.4 & 39.8 & 66.6 & 68.5 & 67.0 & 47.5 \\
UniNER-13B & 13B & 48.7 & 36.2 & 54.2 & 60.9 & 64.5 & 61.4 & 63.5 & 55.6 \\
GoLLIE & 7B & 63.0 & 43.4 & 59.1 & 62.7 & 67.8 & 57.2 & 55.5 & 58.0 \\
\midrule
\emph{few-shot} & & & & & & & & & \\
GPT-4o & - & 60.12 & 55.28 & 49.98 & 76.79 & 68.71 & 53.98 & 75.85 & 62.95 \\
\midrule
\emph{instruction tuned} & & & & & & & & & \\
\emph{or finetuned} & & & & & & & & & \\
RA-IT (50K) & 8B & 45.18 & 40.78 & 58.01 & 63.60 & 64.76 & 61.90 & 62.79 & 56.72 \\
GLiNER-L & 450M & 57.2 & 42.9 & 57.2 & 64.2 & 69.6 & 72.6 & 62.6 & 60.9 \\
GNER-LLaMA & 13B & 68.6 & 47.5 & 63.1 & 68.2 & 75.7 & 69.4 & 69.9 & 66.1 \\
GNER-T5 & 770M & 62.5 & 51.0 & 68.2 & 68.7 & 81.2 & 75.1 & 76.7 & 69.1 \\
SLIMER & 7B & 50.9 & 38.2 & 50.1 & 58.7 & 60.0 & 63.9 & 56.3 & 54.0 \\
\midrule
\emph{ours- few-shot} & & & & & & & & & \\
\textbf{LOCUS} & 470M & 78.04 & 67.81 & 62.88 & 62.9 & 74.37 & 76.39 & 65.99 & \textbf{69.76} \\
\textbf{LOCUS$_\text{mini}$} & \textbf{5M} & 77.80 & 66.14 & 60.36 & 63.86 & 74.31 & 76.07 & 64.40 & 69 \\
\bottomrule
\end{tabular}
\caption{Performance of LOCUS and LOCUS$_\text{mini}$ on MIT \{Movie, Restaurant\} and CrossNER datasets \{AI, Literature, Music, Politics, Science\} as well as the performance of other baselines. LOCUS numbers are average of three individual runs. Extensive baselines can be found in Appendix \ref{sec:appendixC}}
\label{tab:combined_ner_results}
\end{table*}

\section{Related works}
\label{sec:related-works}
The few-shot learning method used in GPT-4 \cite{brown2020language} can be effective for tasks such as NER and TC. However, it is expensive to run, and the performance could be unstable. Two recent frameworks specialize in customizing models and providing effective data generation: DataTune \cite{gandhi2024better}, which modifies publicly available datasets to meet changing needs, and Prompt2Model \cite{viswanathan2023prompt2model}, which synthesizes task-specific examples based on few-shot data. In both approaches, large language models (LLMs) are used to create or transform training sets. LOCUS not only leverages few-shot data but also ensures robust performance across tasks through efficient data generation and fine tuning. By systematically configuring LoRA adapters and generating balanced data from broad sources, we combine the strengths of few-shot generation while achieving a unique balance of efficiency and domain coverage.
\begin{figure}[ht!]
\centering
\includegraphics[width=1\linewidth]{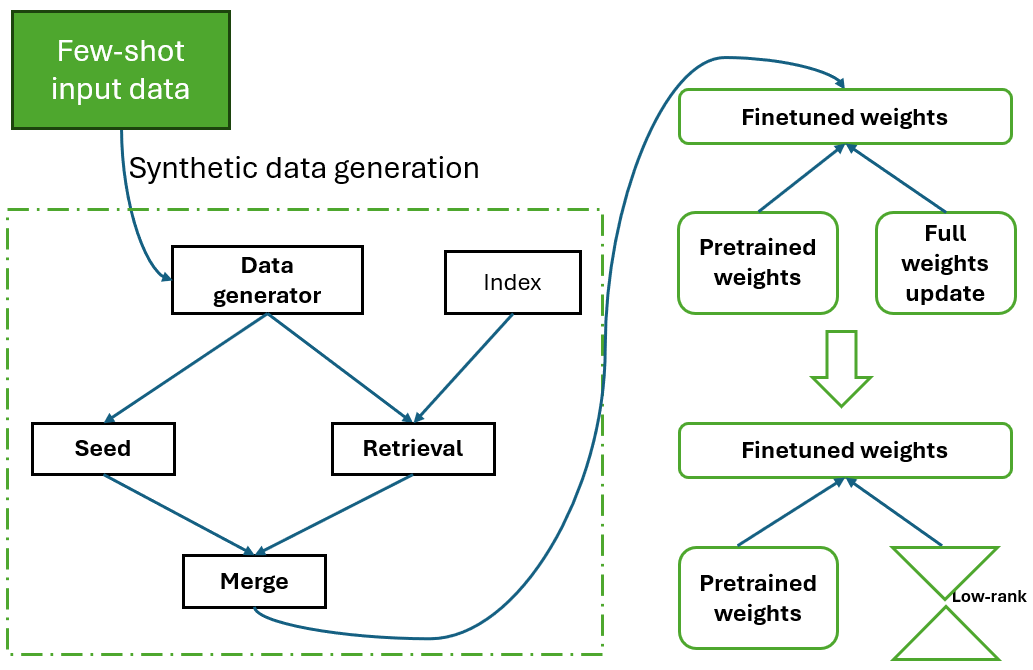}
\caption{LOCUS overview}
\label{fig:main_figure}
\end{figure}

\section{LOCUS overview}
\begin{algorithm}[hbt!]
\small
\caption{LOCUS: Low Cost Universal Specialization}
\label{alg:locus}
\begin{algorithmic}[1]
\REQUIRE $D_{user}$: Small user-labeled dataset  (e.g., 10 per label); $n$: seed count; $\lvert A\rvert$: synthetic size ; $k$: retrieval param ; $s$: final set size .
\STATE \textbf{Seed Selection:} Randomly sample $n$ items from $D_{user}$ to yield seed set $S$.
\STATE \textbf{Seed-Based Generation:} (1) Construct an LLM prompt to generate realistic, diverse synthetic data. (2) Use an LLM to create synthetic samples, forming set $A$.
\STATE \textbf{Iterative Retrieval Generation (repeat $R$ times):}
\begin{enumerate}[label*=\arabic*, itemsep=0pt, topsep=0pt, parsep=0pt, partopsep=0pt]
\item Retrieve top-$k$ examples relevant to $S$ via embeddings.
\item From these, select $m$ for prompt context.
\item Generate $s$ new examples via LLM; append them to $B$.
\end{enumerate}
\STATE \textbf{Combine Data:} $X = A \cup B$.
\STATE \textbf{Fine-Tune Model:} Train a target NER/TC model (e.g., DeBERTa v3) on $X$.
\STATE \textbf{LoRA Decomposition:} Decompose the trained model with LoRA.
\STATE \textbf{Return:} The expanded dataset $X$ and LoRA-based specialized model.
\end{algorithmic}
\end{algorithm}

\noindent Figure \ref{fig:main_figure} illustrates the overall workflow of (\textbf{LOCUS}) framework, wherein a small amount of few-shot labeled data is expanded through both synthetic and retrieved examples before customizing either a NER or TC model. The complete procedure is summarized in Algorithm \ref{alg:locus}. 

\paragraph{Few-Shot Input Data} LOCUS begins with a small user-labeled dataset $D_{user}$, typically containing only a handful of examples per label (around 10). As shown in Algorithm \ref{alg:locus} (\textit{steps} 1 and 2), we randomly select a \emph{seed set} $S$ of size $n$ from $D_{user}$, forming the core examples that will guide all subsequent data augmentation and retrieval stages.

\subsection{Synthetic Data Generation} LOCUS employs two complementary strategies: seed-based and retrieval-based generation for efficient dataset creation. 

\paragraph{Seed-Based Generation} Following \textit{step} 2 of Algorithm \ref{alg:locus}, we craft a meta-prompt by drawing on $A$ (refer Appendix \ref{sec:appendixD}). This prompt reflects the target label schema and domain constraints. An LLM (e.g., GPT-4o, extendbale to others) produces synthetic samples, forming set $A$. This approach ensures that the expanded data mirrors the linguistic diversity and entity definitions in $D_{user}$.

\paragraph{Iterative-Retrieval based Generation} Building on the initial synthetic batch, \textit{step} 3 of Algorithm \ref{alg:locus} iterates retrieval and generation. First, embedding-based retrieval locates the top-$k$ most similar examples from a broader corpus. A subset of these retrieved examples is injected into an LLM prompt as retrieved examples, guiding the production of $s$ new synthetic items. Repeating this cycle accumulates a diverse set $B$ that combines domain fidelity with real-world variations, as illustrated by the “Retrieval” and “Merge” blocks in Figure \ref{fig:main_figure}.

\subsection{Model Specialization: Full vs. LoRA}
\textit{Steps} 5 and 6 of Algorithm \ref{alg:locus}  address how $X$ is used to fine-tune a base model for tasks such as named entity recognition (NER) or text classification (TC). Two pathways are presented:

\paragraph{Full Weights Update}
LOCUS in its conventional form (\textbf{full fine-tuning}) updates all model parameters. Although more resource-intensive, this approach maximizes performance when sufficient computational resources are available.

\paragraph{Low-Rank Adapters (LoRA)}
For users constrained by limited hardware, LOCUS$_{mini}$ deploys (LoRA) to fine-tune only a subset of the parameters. This parameter-efficient strategy significantly reduces memory usage while preserving competitive accuracy, as shown in the “low-rank” branch of Figure \ref{fig:main_figure}.


\section{Experiments}
In this section, we evaluate LOCUS on several NER and TC benchmarks

\subsection{Datasets}

\label{sec:ner}
We assess LOCUS for NER tasks using the MIT \cite{liu2013asgard} and CrossNER \cite{liu2021crossner} datasets, and on ATIS \cite{hemphill1990atis}, AGNews, Yahoo, DBPedia \cite{zhang2015character} for TC benchmarks.

\subsection{Settings}
\begin{figure}[t] 
\centering
\includegraphics[width=1.04\linewidth]{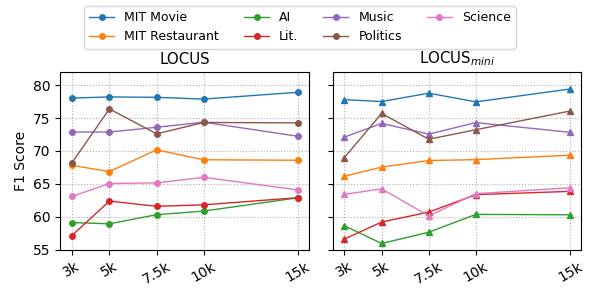}
\caption{An ablation study illustrating how F1 scores change for LOCUS and LOCUS$_\text{mini}$ as dataset size increases. Seven datasets are plotted in different colors.}
\label{fig:locus_ablation}
\end{figure}

\paragraph{Training details} All LOCUS runs utilized the \emph{DeBERTa V3} base model \cite{he2021debertav3} for TC and \emph{mDeBERTa} for NER tasks, which features 12 layers and a 768-dimensional hidden size. 
Synthetic examples were generated via GPT-4o, guided by the Universal-NER pile dataset \cite{zhou2023universalner} as the retrieval index (Figure \ref{fig:main_figure}. Low-Rank Adapters (LoRA) were incorporated with rank and alpha set to 32 (refer Appendix \ref{sec:appendixA}), and the model was trained for 40 epochs at an initial learning rate of $2*10^{-5}$, employing early stopping with a patience of 5. Under these conditions, the LoRA-based checkpoint required is roughly 20 MB (compared to 1.75GB for the baseline model for NER). As part of data preparation, we randomly select ten instances of each entity/class from the corpora to form the seed set. For TC, each dataset has roughly 3k generated examples and for NER, we choose the dataset size based on performance (Figure \ref{fig:locus_ablation}).

\subsection{Results.}
Table \ref{tab:combined_ner_results} shows the performance of our LOCUS models on NER benchmarks while also presenting several zero-shot, few-shot, and fine-tuned baselines.

\textbf{GPT-4o} is used in a few-shot fashion with prompting, it excels in general reasoning but not specifically fine-tuned for NER. It achieves strong cross-domain accuracy, yet may be outperformed by specialized NER models. GPT (3.5 turbo and 4o) are pretrained baselines used in zero‐ and few‐shot modes, respectively, without specific tuning. 
\subsubsection{Baselines}
UniNER‐7B attempt to bridge zero‐shot gaps via conversation‐style or definition‐based instruction tuning, yet can encounter high label overlap between training and test sets. RA‐IT (50K) \cite{xie2024retrieval} exploits retrieval‐augmented data, while InstructUIE \cite{wang2023instructuie}, GNER‐LLaMA, and GoLLIE \cite{sainz2023gollie} leverage instruction or code‐style tuning. Despite smaller backbones or specialized schemes, some of these models still rely on extensive training data, overlapping labels, or carefully designed prompts. SLIMER \cite{zamai2024show} (with or without definitions/guidelines) alleviates confusion for new tags in a zero‐shot manner, but can remain sensitive to prompt clarity.

For TC, in Table \ref{tab:locusTC}, PESCO reframes text classification as a prompt-based text matching with label retrieval and a self-training contrastive loop. CBU identifies instability in GPT-3’s few-shot performance due to inherent biases in prompt construction and propose a contextual calibration method while WC-SBERT propose a zero-shot text classification approach with a novel label-based self-training strategy. 

LOCUS models stand out by operating under few‐shot conditions with minimal train/test overlap and uses far less memory than conventional fine‐tuning. Despite this efficiency, our approach outperforms far larger LLMs on both zero‐/few‐shot and surpasses/matches most instruction‐tuned models trained on substantially more data. Consequently, we offer a balanced solution that simultaneously handles novel labels, uses low resource, and achieves top‐tier F1 scores across benchmarks.

Additionally, we also show that LOCUS surpasses few‐shot GPT‐4o baseline on MultiNERD (refer Appendix \ref{sec:appendixB}), achieving an overall average of 70.77 versus 68.55.

\begin{table}[!h]
\centering
\small
\begin{tabular}{l c c c c c}
\toprule
\textbf{Model} & \textbf{\textit{param}} & \textbf{Atis} & \textbf{AgNews} & \textbf{Yahoo} & \textbf{DbPedia} \\
\midrule
\textbf{Claude} & - & - & 82.6 & 68.0 & - \\
\textbf{GPT-4o} & - & 90 & 89.1 & 74.3 & 99.1 \\
\midrule
\textbf{PET} & 355M & -- & 79.4 & 56.4 & 75.2 \\
\textbf{PESCO} & 340M & -- & 89.6 & 71.1 & 98.5 \\
\textbf{CBU} & 175B & -- & 84.3 & -- & 86.9 \\
\textbf{WC-SBERT} & 110M & & 81.5 & 62.7 & 74.81 \\
\textbf{LOCUS} & 185M & 91.4 & 87.8 & 67.6 & 96.1 \\
\textbf{LOCUS$_{mini}$} & 5M & 86.4 & 87.3 & 67.5 & 98.2 \\
\bottomrule
\end{tabular}
\caption{LOCUS and LOCUS$_{mini}$ performance on TC datasets - \{Atis, AgNews, Yahoo, DBPedia\} comapred with baselines PET \cite{schick2020exploiting}, PESCO \cite{wang2023pesco}, CBU \cite{zhao2021calibrate}, WC-SBERT \cite{chi2023wc}}
\label{tab:locusTC}
\end{table}

\section{Conclusion}
In summary, we introduced LOCUS, a cost‐efficient pipeline that unifies data retrieval, synthetic generation, and parameter‐efficient fine‐tuning for specialized NLP tasks. Comprehensive experiments show notable performance gains over large LLMs like GPT‐4o, while drastically reduced memory requirements. LOCUS requires minimal few-shot data, while balancing accuracy, resource consumption, and adaptability, making specialized NLP models more accessible.

\section*{Limitations}
LOCUS relies on a large universal dataset for retrieval even if it operates on few-shot inputs. The universal dataset may produce few matching sentences in situations involving highly specialized fields or low-resource languages, which could result in subpar retrieval. Incomplete few-shot data may introduce noise through synthetic synthesis. Lastly, while the performance of LOCUS$_{mini}$ is strong, it is not assured to be on par with completely optimized performance across all domains.


\bibliography{custom}

\pagebreak
\pagebreak
\appendix

\section{LoRA ablation experiments}
\label{sec:appendixA}
\begin{figure}[t] 
\centering
\includegraphics[width=1.04\linewidth]{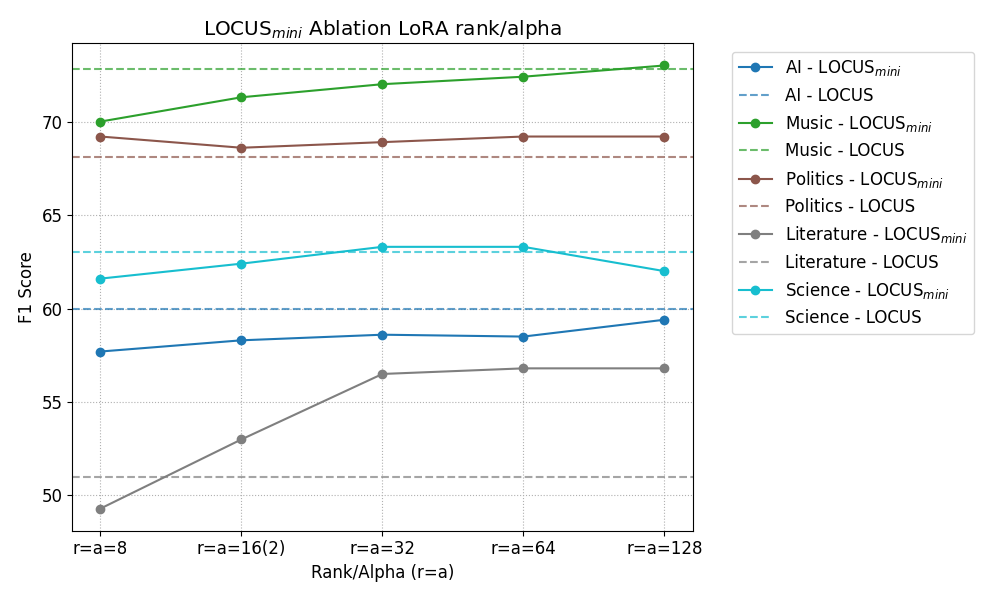}
\caption{An ablation study illustrating how F1 scores change for LOCUS$_\text{mini}$ as we increase rank/alpha compared with LOCUS.}
\label{fig:lora_ablation}
\end{figure}

Figure \ref{fig:lora_ablation} shows the performance of LOCUS and it's mini version as a function of LoRA rank and alpha.


\section{Experiments on MultiNERD datasets}
\label{sec:appendixB}
Table \ref{tab:multinerd-results} shows the performance of LOCUS on MultiNERD dataset.  LOCUS or LOCUS$_{mini}$ obtains better F1 in most languages, reaffirming its consistent multi‐lingual strength in few‐shot setups. 

\begin{table}[hbt!]
\centering
\begin{tabular}{lccc}
\hline
\textbf{dataset} & \textbf{LOCUS} & \textbf{LOCUS$_{mini}$} & \textbf{GPT-4o} \\ \hline
EN  & 71.66 & 71.84 & 70.19 \\
ES  & 70.98 & 71.61 & 70.34 \\
DE  & 69.99 & 68.32 & 70.09 \\
IT  & 69.77 & 70.62 & 67.17 \\
PL  & 69.82 & 66.21 & 66.06 \\
PT  & 71.43 & 70.17 & 65.93 \\
FR  & 72.18 & 72.33 & 70.07 \\ \hline
AVG & \textbf{70.77} & 69.33 & 68.55 \\ \hline
\end{tabular}
\caption{Performance of LOCUS on MultiNERD Datasets against few-shot GPT-4o}
\label{tab:multinerd-results}
\end{table}

\section{Performance of LOCUS and LOCUS$_{mini}$}
\label{sec:appendixC}

Table \ref{tab:combined_ner_results_extended} shows extensive baselines compared to LOCUS on MIT and CrossNER benchmarks.

\begin{table*}[hbt!]
\centering
\begin{tabular}{l c c c c c c c c c}
\toprule
\textbf{Model} & \textbf{\textit{params}} & \multicolumn{2}{c}{\textbf{MIT}} & \multicolumn{5}{c}{\textbf{CrossNER}} & \multicolumn{1}{c}{\textbf{AVG}} \\
\cmidrule(lr){3-4}\cmidrule(lr){5-9}
& & \textbf{Movie} & \textbf{Res.} & \textbf{AI} & \textbf{Lit.} & \textbf{Music} & \textbf{Politics} & \textbf{Science} & \\
\midrule
\emph{zero-shot} & & & & & & & & & \\
Vicuna-7B & 7B & 6.0 & 5.3 & 12.8 & 16.1 & 17.0 & 20.5 & 13.0 & 13.0 \\
Vicuna-13B & 13B & 0.9 & 0.4 & 22.7 & 22.7 & 26.6 & 27.0 & 22.0 & 17.5 \\
USM & 2B & 37.7 & 17.7 & 28.2 & 56.0 & 44.9 & 36.1 & 44.0 & 37.8 \\
InstructUIE & 11B & 63.0 & 21.0 & 49.0 & 47.2 & 53.2 & 48.1 & 49.2 & 47.2 \\
GPT-3.5-turbo & 175B & 5.3 & 32.8 & 52.4 & 39.8 & 66.6 & 68.5 & 67.0 & 47.5 \\
UniNER-7B & 7B & 42.4 & 31.7 & 53.5 & 59.4 & 65.0 & 60.8 & 61.1 & 53.4 \\
UniNER-13B & 13B & 48.7 & 36.2 & 54.2 & 60.9 & 64.5 & 61.4 & 63.5 & 55.6 \\
GoLLIE & 7B & 63.0 & 43.4 & 59.1 & 62.7 & 67.8 & 57.2 & 55.5 & 58.0 \\
\midrule
\emph{few-shot} & & & & & & & & & \\
GPT-4o & - & 60.12 & 55.28 & 49.98 & 76.79 & 68.71 & 53.98 & 75.85 & 62.95 \\
\midrule
\emph{instruction tuned} & & & & & & & & & \\
\emph{or finetuned} & & & & & & & & & \\
RA-IT (50K) & 8B & 45.18 & 40.78 & 58.01 & 63.60 & 64.76 & 61.90 & 62.79 & 56.72 \\
UniNER-type+sup & 7B & 61.2 & 35.2 & 62.9 & 64.9 & 70.0 & 66.9 & 70.8 & 55.4 \\
GoLLIE & 13B & 63.0 & 43.4 & 59.1 & 62.7 & 67.8 & 57.2 & 55.5 & 58.4 \\
GLiNER-L & 450M & 57.2 & 42.9 & 57.2 & 64.2 & 69.6 & 72.6 & 62.6 & 60.9 \\
GNER-LLaMA & 13B & 68.6 & 47.5 & 63.1 & 68.2 & 75.7 & 69.4 & 69.9 & 66.1 \\
GNER-T5 & 770M & 62.5 & 51.0 & 68.2 & 68.7 & 81.2 & 75.1 & 76.7 & 69.1 \\
SLIMER & 7B & 50.9 & 38.2 & 50.1 & 58.7 & 60.0 & 63.9 & 56.3 & 54.0 \\
\midrule
\emph{ours- few-shot} & & & & & & & & & \\
\textbf{LOCUS} & 470M & 78.04 & 67.81 & 62.88 & 62.9 & 74.37 & 76.39 & 65.99 & \textbf{69.76} \\
\textbf{LOCUS$_{mini}$} & \textbf{5M} & 77.80 & 66.14 & 60.36 & 63.86 & 74.31 & 76.07 & 64.40 & 69 \\
\bottomrule
\end{tabular}
\caption{Performance of LOCUS and LOCUS$_{mini}$ on MIT \{Movie, Restaurant\} and CrossNER datasets \{AI, Literature, Music, Politics, Science\} as well as the performance on extensive baselines. LOCUS numbers are average of three individual runs.}
\label{tab:combined_ner_results_extended}
\end{table*}

\section{Meta Prompts for Data Generation}
Figure \ref{fig:prompts} shows a concise version of meta prompts used for seed-based and retrieval-based generation for NER. 
\label{sec:appendixD}
\begin{figure}[p]
\raggedright

Seedbased Generation Meta Prompt for NER

Generate a diverse dataset with features: \\
(1) Real-world entities  \\
(2) Multifactorial, including outliers for robustness. \\
(3) Domain: \textit{target\_domain}, Entities: \textit{target\_entities}, guided by \textit{entity\_examples}. \\

Return format:
Text: [text greater than a certain length] \\
Entities: [among the given entity list] \\
----------------\\
 
Retrieval-Based Generation Meta Prompt for NER

Given a context of relevant samples, generate diverse data: \\
(1) Real-world entities \\
(2) Multifactorial with outliers. \\
(3) Domain: \textit{target\_domain}, Entities: {target\_entities}, guided by \textit{entity\_examples} with \textit{retrieved\_examples}\\

Return format:
Text: [text greater than a certain length] \\
Entities: [among the given entity list]

\caption{Concise prompts for seed-based and retrieval-based generation}
\label{fig:prompts}
\end{figure}

\end{document}